# Predictive Accuracy of a Hybrid Generalized Long Memory Model for Short Term Electricity Price Forecasting

Souhir Ben Amor[1,2] Heni Boubaker[3] Lotfi Belkacem[4]

**Abstract**

Accurate electricity price forecasting is the main management goal for market participants since it represents the fundamental basis to maximize the profits for market players. However, electricity is a non-storable commodity and the electricity prices are affected by some social and natural factors that make the price forecasting a challenging task. This study investigates the predictive performance of a new hybrid model based on Generalized long memory autoregressive model (k-factor GARMA), the Gegenbauer Generalized Autoregressive Conditional Heteroscedasticity(G-GARCH) process, Wavelet decomposition, and Local Linear Wavelet Neural Network (LLWNN) optimized using two different learning algorithms; the Back propagation algorithm (BP) and the Particle Swarm optimization algorithm (PSO). The performance of the proposed model is evaluated using data from Nord Pool Electricity markets. Moreover, it is compared with some other parametric and non-parametric models in order to prove its robustness. The empirical results prove that the proposed method performs well than other competing techniques.

**Keywords:** Wavelet decomposition, WLLWNN, *k*-factor GARMA, G-GARCH, Electricity spot price, Forecasting.

**Research highlights**

- The wavelet decomposition is used to improve the forecasting accuracy of the LLWNN model.
- A hybrid method is developed to forecast the electricity price.
- The hybrid method combines the parametric k-factor GARMA model, Wavelet decomposition, and the non-parametric LLWNN methods.
- The learning algorithms; BP and PSO methods are applied to obtain optimum WLLWNN structure to avoid overfitting.

---

[1] Institute of high commercial studies of Sousse (IHEC), 40 Route de ceinture Sahloul III, 4054 Sousse, Tunisia
[2] Corresponding author.
[3] Institute of high commercial studies of Sousse (IHEC
[4] Institute of high commercial studies of Sousse (IHEC)

- The Dual generalized k-factor GARMA-G-GARCH model is adopted for comparison purpose.

**Nomenclature**

$\phi$, Father wavelet; $\psi$, Mother wavelet; $\varphi(x)$, is the active wavelet functions; $X(t)$, a linear combination at arbitrary level $J_0 \in N$ through different scales; $s_{J_0,k}$ and $d_{j,k}$ are the detail (fine scale) coefficients; $h_l$, $l=0,\ldots,L-1$ is the high-pass wavelet filter; $g_l$, $l=0,\ldots,L-1$ is the low pass scaling filter; $E$, The objective function to minimize of the Back propagation (BP) algorithm; $\omega_{i,j}$, represent the connection weight; $p$, is the number of input $(i=1,2,....p)$ and $l$ is the number of the hidden units $(j=1,2,....l)$ ; $r$, is the learning rate of the BP algorithm; $e$, Error between output values $\hat{y}$ by and real values $y$ ; $pbest$, the best value of each agent (in the PSO algorithm); $gbest$, the best Value of each agent in the group; $v_i$, is the velocity of agent $i$; $s_i$, is the current position of agent $i$; $a_i$ and $b_i$ are the scale and translation parameters, respectively, associated to the mother wavelet $\psi$; $\upsilon_i$, is the linear model associated to the LLWNN weights; $Y$, the output of the LLWNN model; $\lambda_m$, Gegenbauer frequency of the *k*-factor GARMA model; $\Phi(L)$ and $\Theta(L)$, Polynomials of the delay operator $L$, related to the *k*-factor GARMA model; $d_m$, Long-memory parameter associated with the *k*-factor GARMA model; $P_m(L)$,Gegenbauer polynomial of the *k*-factor GARMA model; $\mu_t$, the conditional mean of the time series; $\varepsilon_t$, the residuals at time $t$ from the $k$-factor GARMA model; $f$, is a non-linear, non-parametric function determined by the neural network; $\sigma_t^2$, is the conditional variance; $\lambda_v$, Gegenbauer frequency of the *k*-factor GARMA model; $P_v(L)$, Gegenbauer polynomial of the G- GARCH model; N, is the number of observations of our sample.

## 1. Introduction

Since early 80s, there has been a radical change in the electricity market all across the world. The electricity market, which was largely functioning in a vertically integrated operating environment (where generation, transmission and distribution combined in one hand) shifted to a system where all these activities got separated and independent. The increasing privatization and deregulation of the market have also led to sharper competition (Amjady (2006)).

In power markets, price investigation has become an essential topic for all its participants. Background information about the electricity price is important for risk management. In fact, it may stand an advantage for a market player facing competition. Moreover, forecasting electricity prices at different time horizons is valuable for all industry stakeholders for cash flow analysis, capital budgeting, integrated resource planning, regulatory rule-making and financial procurement. Therefore, producers and consumers rely on price forecasting information to propose their corresponding bidding strategies.

However, the electricity prices has a behavior that is different from that of other commodities. The main difference is that electricity is a non-storable merchandize, accordingly a small variation in electricity production in hours or minutes can lead to huge price changes. Moreover, electricity prices show some particular characteristics such as high frequency, non-stationary behavior, multiple seasonality (on annual, weekly, and daily levels), calendar effect, high volatility, high percentage of unusual prices, hard nonlinear behavior, long memory, mean reversion, price spikes and limited information to the market participants, which may affect the prices dramatically (Weron (2006). In this framework, due to the complexity of the electricity market, price forecasting has been the most challenging task. This has also motivated the researchers to develop efficient and intelligent methods to forecast the prices so that all stakeholders in the market can benefit out of it.

Herby, an appropriate forecast model for electricity prices should consider the previous mentioned features. Meanwhile, various methods from different modeling families are used for different analysis or planning issues. In the field of electricity price forecasting, broadly two methods; statistical or econometric time series models and soft computing models, these two approaches are found to have been applied.

In statistical models, Auto regressive integrated moving average ARIMA (Contreras et al. (2003)), and Generalized Autoregressive Conditional Heteroscedasticity GARCH (Garcia et al. (2005); Ghosh and Kanjilal (2014) and Girish. (2016)) are used extensively. Some authors combine the ARIMA and GARCH models to take into consideration the mean and variance of the time series when modeling and forecasting the electricity price (Bowden and Payne (2008); and Heping and Jing (2013)). However, these models does not allow to take into consideration the long memory behavior that characterize the electricity prices. To overcome this limitation, some authors applied the Fractional Auto Regressive Moving Average (FARMA) model and the Fractionally Integrated Generalized Autoregressive Conditional Heteroscedasticity (FIGARCH) process, to model finite persistence in the conditional mean and the conditional variance, respectively (Koopman et al. (2007), Najeh et al. (2012)). In the spectral domain, these processes have a peak for very low frequencies close to the zero frequency. Hence, it is noteworthy that ARFIMA model is not able to model the persistent periodic or cyclical behavior in the time series. To overcome this insufficiency, Gray et al. (1989) introduced a second generation of the long-memory model termed a generalized (seasonal) long-memory or Gegenbauer Autoregressive Moving Average (GARMA) process, which has been established to estimate both the seasonality and the persistence in the data. Woodward et al. (1998) generalizes the single frequency GARMA process to the so-called $k$-factor GARMA process that permits the spectral density function to be not just located at a single frequency but presented a finite number $k$ of frequencies in $[0,\pi]$, identified as the Gegenbauer frequencies or G-frequencies. The $k$-factor GARMA model has been applied by several authors to reproduce the seasonal patterns as well as the persistent effects in the stock markets (Boubaker and Sghaier (2015), Caporale and Gil-Alana (2014), Caporale et al. (2012)). Despite the compatibility of this model with the characteristics of electricity prices, few applications are oriented in the electricity market (Diongue et al. (2009), Soares and Souza (2006) and Diongue et al. (2004)).

However, the $k$-factor GARMA model allows only to model the conditional mean, assuming that the residuals are white noise with constant variance over time. However, this assumption is not proved in practice and the residuals are often characterized by a time-varying variance.

To overcome this limitation, some researchers suggested to extend the $k$-factor GARMA model using the GARCH, FIGARCH and Fractionally Integrated Asymmetric Power ARCH (FIAPARCH) models (Boubaker (2015) Boubaker and Boutahar (2011) Boubaker and Sghaier (2015)). Nevertheless, these models are not fully satisfactory in the modelling the volatility of intra-daily financial time series. The main feature of such data is the strong evidence of cyclical patterns in the volatility. The empirical evidence highlights the importance of modelling the periodic dynamics of the volatility. For this aim, Bordignon et al. (2007, 2010) suggested new category of GARCH models characterized by periodic long memory behavior. This category of models introduces Gegenbauer polynomials into the equation of standard GARCH model, In the literature, only few studies apply this process for modelling the electricity spot price (Diongue et al. (2009)).

For the AI methods, application of artificial neural network (ANN) for the electricity price forecasting which considered as generalized periodic long-memory filters to estimate the time-varying volatility has become a research hotspots (Anbazhagan and Kumarappan (2014), Ioannis and Athanasios (2016), Harmanjot et al. (2016); and Jesus et al (2018)). Zhang and Benveniste (1992) suggested the Wavelet Neural Networks, which adopt a wavelet as an activation function, as an alternative to the classic sigmoidal functions. The WNs have been effectively used in the electricity prices forecasting in the short-term, (Bashir and El-Hawary (2000); Yao, Song, Zhang, and Cheng (2000); Gao and Tsoukalas (2001); Benaouda, et al. (2006); Ulugammai et al. (2007); Pindoriya et al (2009); and Mashud and Irena (2016)). Chen et al. (2004) developed a new type of wavelet neural network termed the Local Linear Wavelet Neural Network (LLWNN). Several researchers for the electricity price forecasting have extensively used the LLWNN model (Pany (2011); Chakravarty et al. (2012); Pany et al. (2013); and Athanassios et al. (2015)).

it is worth noting that a time series is often complicate in nature and an individual model is not able to detect the different patterns in the same way, in such case, adopting a hybrid methods or combining several models (Granger (1989)) has become a common practice to overcome the limitations of using a single model and enhance the forecasting accuracy.

In the literature, several combination methods have been suggested (Tseng et al. (2002); Zhang (2003); Taskaya and Casey (2005); Valenzuela et al. (2008); Khashei and Bijari (2010); Tan et al. (2010) and Jiang et al. (2017)).

Some researchers suggested that this combination can be improved using the wavelet decomposition. To exemplify, Shafie-khah et al. (2011) developed a novel hybrid method to forecast day ahead electricity price. This hybrid method is based on wavelet transform, Auto-Regressive Integrated Moving Average (ARIMA) models and Radial Basis Function Neural Networks (RBFN). Zhang et al (2017) proposed a hybrid approach that combines the wavelet transform, the kernel extreme learning machine (KELM) based on self-adapting particle swarm optimization and an auto regressive moving average (ARMA). Jinliang et al.(2018) proposed a new hybrid model based on improved empirical mode decomposition (IEMD), autoregressive integrated moving average (ARIMA) and wavelet neural network (WNN) for Short term electricity load forecasting.

Our research focuses on resolving the issues of modeling and forecasting the feature of the electricity prices. In this framework, this paper provides three contributions. The main one is to improve the forecasting accuracy of LLWNN model. This objective is achieved through using the wavelet theory to decompose the historical price, instead of introducing it directly to the Network (in the input layer) and assess the effect of different levels of decomposition on forecasting accuracy. This technique allows the network to detect the existence of seasonal long memory behavior and thus better estimate the data, this novel network is termed WLLWNN. In fact, previous studies combined the wavelet decomposition with the ANN (Aggarwal et al. (2008)), this technique proves its performance but it's not applied with the LLWNN. Meanwhile, some research (Pany (2011), Pany and Ghoshal (2013), Chakravarty et al. (2012)) assumed that this model can provide prediction accuracy since the characteristics associated to the electricity prices can be identified in the hidden layer by means of the wavelet activation function without needing any external decomposer/ composer.

Secondly, a new hybrid model is proposed in order to exploit the strength of both parametric and non-parametric approaches, this model is called $k$-factor GARMA-WLLWNN process, which allows for long-memory behavior, associated with $k$-frequency and include a WLLWNN type model to describe time-varying volatility.

Thirdly, in the literature of generalized long memory models, the authors adopt either the $k$-factor GARMA model or the G-GARCH model to estimate the conditional mean and the conditional variance of the time series, respectively. However, none of them considers the long memory and cyclical behavior simultaneously in the conditional mean and the conditional variance. In this study, with the aim of providing a robust tool for electricity prices prediction, we develop a new approach based on dual generalized long memory process using both the $k$-factor GARMA model and the G-GARCH model, which allows taking into account many stylized facts observed on the electricity spot prices, in particular stochastic volatility, long-rang dependence and multi-seasonality behaviors.

Moreover, through this study we aims at proving the performance of the proposed $k$-factor GARMA-WLLWNN model by comparing it with the $k$-factor GARMA-G-GARCH model. In fact, this step consist of testing the performance of the novel WLLWNN with a parametric model (G-GARCH) in modeling and forecasting the periodic long memory behavior that existing in the conditional variance.

The log-return of electricity price for the Nord Pool market is used in order to show the appropriateness and effectiveness of the proposed model to time series forecasting. The rest of the paper is organized as follows; in the next section, we present the econometric methodology, which includes the theoretical concepts of wavelet decomposition, the learning algorithms for optimizing the neural networks and the basic concept of the proposed hybrid $k$-factor GARMA-WLLWNN model and the $k$-factor GARMA-G-GARCH model. Section 3 deals with the empirical framework, where the proposed hybrid model is applied to log-return of electricity spot price forecasting and its performance is compared with the individual LLWNN as well as the individual WLLWNN model, the hybrid $k$-factor GARMA-LLWNN model and the generalized long memory $k$-factor GARMA-G-GARCH model, and section 4 concludes the paper.

## 2. Econometric Methodology

### 2.1 Theoretical concepts of wavelet decomposition

Electricity price series exhibits specific features and riche structure; for this reason, signal-processing techniques like Fourier Transfer and wavelets are good candidates for bringing out hidden patterns in price series (Nicolaisen et al. (2000)). In order to tackle the problem related to the specific behavior of electricity price series, wavelets have been adopted since they can produce a good local representation of the signal in both frequency and time domains. Wavelet decomposition is applied for multi-scale analysis of the signal and decomposes the time series signal into one low-frequency sub-series (constitute the approximation part) and some high-frequency sub-series (constitute the detailed part) in the wavelet domain. These constitutive series have improved statistical properties than original price series and consequently, improved forecasting accuracy can be achieved by their suitable utilization.

Wavelets are used to analyze the time series through mathematical rules, where the time series is decomposed on a scale-by-scale basis. Hence, using a dilation and translation operations, this technique allow a flexible time-frequency resolution, and can define local features of a given function in a parsimonious way. Wavelets are orthonormal bases attained through dyadically dilating and translating a pair of specially constructed functions denotes by $\phi$ and $\psi$, which are named father wavelet and mother wavelet, respectively, given by:

$$\int \phi(t)dt = 1, \tag{1}$$

$$\int \psi(t)dt = 0. \tag{2}$$

The smooth and the low-frequency part of the time series are detected by means of the father wavelet while the detail and the high-frequency components are defined by the mother wavelet. The obtained wavelet basis is:

$$\phi_{j,k}(t) = 2^{j/2}\phi\left(2^{j}t - k\right), \quad (3)$$

$$\psi_{j,k}(t) = 2^{j/2}\psi\left(2^{j}t - k\right). \quad (4)$$

Where $j = 1,\ldots,J$ indexes the scale and $k = 1,\ldots,2^{j}$ indexes the translation. The parameter $j$ is adopted as the dilation parameter of the wave's functions. This parameter $j$ adjusts the support of $\psi_{j,k}(t)$ in order to locally detect the features of high or low frequencies. The parameter $k$ is employed to relocate the wavelets in the temporal scale. The number of observations limits the maximum number of scales that can be used in the analysis $\left(T \geq 2^{J}\right)$.

The localization property is a special property of the wavelet expansion, where the coefficient of $\psi_{j,k}(t)$ reveals information content of the function at approximate location $k2^{-j}$ and frequency $2^{-j}$. By means of wavelets, any function in $L^{2}(\Re)$ can be extended over the wavelet basis, exceptionally, as a linear combination at arbitrary level $J_0 \in \mathrm{N}$ through different scales of the type:

$$X(t) = \sum_{k} s_{J_0,k}\phi_{J_0,k}(t) + \sum_{j \geq J}\sum_{k} d_{j,k}\psi_{j,k}(t) \quad (5)$$

Where $\phi_{J_0,k}$ a scaling function with the corresponding coarse scale coefficients $s_{J_0,k}$ and $d_{j,k}$ are the detail (fine scale) coefficients given respectively by $s_{J_0,k} = \int X(t)\phi_{J_0,k}(t)dt$ and $d_{j,k} = \int X(t)\psi_{j,k}(t)dt$. These coefficients give a measure of the contribution of the corresponding wavelet to the function. The expression (5) denotes the decomposition of $X(t)$ into orthogonal components at different resolutions and constitutes the wavelet multiresolution analysis (MRA).

In practical applications, we invariably deal with sequences of values indexed by integers rather than functions defined over the entire real axis. Instead of actual wavelets, we use short sequences of values referred to as wavelet filters. The number of values in the sequence is termed the width of the wavelet filter. Hence, the wavelet analysis measured through a filtering perspective is then well suited to time series analysis.

The wavelet coefficients of the discrete wavelet transform, can be considered from the MRA scheme. The recursive MRA scheme[5], which is implemented by a two-channel filter bank (i.e. a high-pass wavelet filter $\{h_l, \quad l = 0, \ldots, L-1\}$[6]and its associated low pass scaling filter $\{g_l, \quad l = 0, \ldots, L-1\}$[7]satisfying the quadrature mirror relationship given by $g_l = (-1)^{l+1} h_{L-1-l}$ for $l = 0, \ldots, L-1$, where $L \in \mathrm{N}$ is the length of the filter) illustration of the wavelet transform, is divided into decomposition and reconstruction schemes referring to the forward and inverse wavelet transform.

Daubechies (1992) has constructed a class of wavelet functions where $\phi$ is a function such that $\{\phi(t-k), \quad k \in Z\}$forms an orthonormal basis of piecewise constant functions of length one. The Daubechies wavelet has many desirable properties; its most useful property that is possessing the smallest support for a given number of vanishing moments.[8] Daubechies defined a useful category of wavelet filters, termed the Daubechies compactly supported wavelet filters of width $L$ and distinguishes between two choices; the extremal phase filters $D(L)$ and the least asymmetric filters $La(L)$.

### 2.2 Learning Algorithms for optimizing the neural networks

#### 2.2.1 The Back propagation Algorithm

For the BP algorithm, at the beginning, the parameters are randomly initialized, and then the algorithm measures the error between the output value and the real value, and finally adjusts the weights in the direction of descendent gradient. The learning rate controls the speed of the training process. If this rate is high, the network will learn quicker, but the learning process will never converge if this rate is too high. In contrast, if the learning rate is so low, the network may converge to a local minimum instead of the global minimum.

---

[5] Mallat's Multiresolution Analysis is considered as a robust theoretical framework for critically sampled wavelet transformation (for more details, see Mallat (1989)).

[6] The wavelet function (filter) of support $L$ proceeds as a special filter possessing specific properties, such that (i) it integrates to zero, i.e., $\sum_{l=0}^{L-1} h_l = 0$,(ii) has unit energy, i.e., $\sum_{l=0}^{L-1} h_l^2 = 1$, and (iii) is orthogonal to its even shifts, i.e., $\sum_{l=0}^{L-1} h_l h_{l+2n} = \sum_{l=-\infty}^{\infty} h_l h_{l+2n} = 0, \quad \forall n \in N^*$.

[7] The scaling filter of support $L$ is defined so as to satisfy the following properties, (i) $\sum_{l=0}^{L-1} g_l = \sqrt{2}$,(ii) $\sum_{l=0}^{L-1} g_l^2 = 1$,(iii) $\sum_{l=0}^{L-1} g_l g_{l+2n} = \sum_{l=-\infty}^{\infty} g_l g_{l+2n} = 0, \quad \forall n \in N^*$.

[8] For Daubechies wavelets, the number of vanishing moments is half the filter length.

The equations of the BP algorithm are presented in detail in (Burton and Harley (1994)) and they are briefly described below.

The objective function to minimize is given as;

$$E = \frac{1}{2}\left[y_t - \omega_{1,0}\varphi_1(x) - \omega_{1,1}p_1\varphi_1(x) - \ldots - \omega_{2,0}\varphi_2(x)\omega_{2,1}p_2\varphi_2(x) - \ldots\omega_{l,0}\varphi_l(x)\omega_{l,1}p_1\varphi_l(x) - \ldots\omega_{l,p}p_p\varphi_l(x)\right] \quad (6)$$

Where $y_t$ is the desired value, $\varphi(x)$ is the active wavelet functions, $\omega_{1,0}$ represent the connection weight, $p$ is the number of input $(i=1,2,\ldots.p)$ and $l$ is the number of the hidden units $(j=1,2,\ldots.l)$. The weight is updated from $i^{th}$ to the $(i+1)^{th}$ iteration, that is from $\omega_t$ to $\omega_{t+1}$ is given by;

$$\omega_{t+1} = \omega_t + \Delta\omega_t = \omega_t + \left(r\frac{\partial E_t}{\partial \omega_t}\right), \quad (7)$$

Denote that $r$ is the learning rate adopted in the WLLWNN model.

Where $\frac{\partial E}{\partial \omega}$ for all weights are described by the following equations;

$$\frac{\partial E}{\partial \omega_{i,0}} = \omega_{i,0} + r.e.\left(\frac{1}{2}\right).\left(x_1^2 + x_2^2 + \ldots + x_p^2\right).\exp(-((x_1 - c_i)^2 + (x_2 - c_i)^2 + \ldots + (x_p - c_i)^2)) \quad (8)$$

For $\forall j \neq 0$;

$$\frac{\partial E}{\partial \omega_{i,j}} = \omega_{i,j} + r.e.\left(\frac{1}{2}\right).\left(x_1^2 + x_2^2 + \ldots + x_n^2\right).\exp(-((x_1 - c_i)^2 + (x_2 - c_i)^2 + \ldots + (x_n - c_i)^2)).x_j \quad (9)$$

That is

$$\frac{\partial E}{\partial \omega_{1,0}} = \omega_{1,0} + r.e.\left(\frac{1}{2}\right).\left(x_1^2 + x_2^2 + \ldots + x_p^2\right).\exp(-((x_1 - c_i)^2 + (x_2 - c_i)^2 + \ldots + (x_p - c_i)^2)) \quad (10)$$

$$\frac{\partial E}{\partial \omega_{1,2}} = \omega_{1,2} + r.e.\left(\frac{1}{2}\right).\left(x_1^2 + x_2^2 + \ldots + x_p^2\right).\exp(-((x_1 - c_i)^2 + (x_2 - c_i)^2 + \ldots + (x_p - c_i)^2)).x_2 \quad (11)$$

$$\frac{\partial E}{\partial \omega_{2,0}} = \omega_{2,0} + r.e.\left(\frac{1}{2}\right).\left(x_1^2 + x_2^2 + \ldots + x_p^2\right).\exp(-((x_1 - c_i)^2 + (x_2 - c_i)^2 + \ldots + (x_p - c_i)^2)) \quad (12)$$

$$\frac{\partial E}{\partial \omega_{2,1}} = \omega_{2.1} + r.e.\left(\frac{1}{2}\right).\left(x_1^2 + x_2^2 + \ldots + x_p^2\right).\exp(-((x_1 - c_i)^2 + (x_2 - c_i)^2 + \ldots + (x_p - c_i)^2)).x_1 \quad (13)$$

Where $e$ is the error between output values $\hat{y}$ by and real values $y$ $\left(e = \hat{y} - y\right)$ ;

The other weights are also updated in the same way.

### 2.2.2 The Particle Swarm optimization Algorithm (PSO)

Kennedy and Eberhart (1995) developed the PSO as an optimization technique. In comparison with other learning algorithms, the PSO proved clearly its efficiency.

PSO algorithm is established through simulation of bird flocking in two-dimension space. The position of each agent is denoted by $XY$ axis position and the velocity is represented by $vx$ and $vy$. The agent position's adjustment is recognised by the position and the velocity information. The Bird flocking optimizes the objective function. Each agent knows its best value so far $(pbest)$ and its $XY$ position. In addition, each agent knows the best value so far in the group $(gbest)$ among $(pbest)$. Mainly each agent tries to adjust its position using the following information.

(a) The distance between current position and $pbest$.

(b) The distance between the current position and $gbest$.

The following equation is adopted to update the Velocity associated to each agent:

$$v_i^{p+1} = wv_i^p + c_1 rand_1 * (pbest_1 - s_i^p) + c_2 rand_2 (gbest - s_i^p) \tag{14}$$

Where $v_i^p$ is the velocity of agent $i$ at iteration $p$, $w$ is the weight function, $c_j$ is weighting factor, $s_i^p$ is the current position of agent $i$ at iteration $p$, $pbest_i$ is the $pbest$ of agent $i$ and $gbest$ is the $gbest$ of the group.

The velocity, which progressively gets close to $pbest$ and $gbest$ can be computed using the above equation. The actual position, which characterises the searching point in the solution space, can be updated using the following equation:

$$s_i^{p+1} = s^p + v_i^{p+1} \tag{15}$$

The first term of equation (14), denote the previous velocity of the agent. The velocity of the agent is updated through the second and third terms.

The general steps, which describe the optimization of the LLWNN using the PSO algorithm can be demonstrated as follows:

*Step.1* The initial condition is generated for each agent:

The initial searching points ($s_i^0$) and velocity ($v_i^0$)of each agent are habitually generated randomly within the allowable range. Note that the dimension of search space contains all the parameters of the LLWNN.

The current searching point is set to $pbest$ for each agent. The best-evaluated value of $pbest$ is set to $gbest$ and the agent number with the best value is stored.

*Step.2* The searching points are evaluated for each agent:

The value of the objective function is calculated for each agent. If this calculated value is improved in comparison with the current $pbest$ of the agent, the $pbest$ value is replaced by the current value. If the best value of $pbest$ is better than the current $gbest$, $gbest$ is replaced by the best value and the agent number corresponding to the best value is stored.

*Step.3* Modification of each searching point:

Using equations (14) and (15), the actual searching point of each agent is updated.

*Step 4*. Verification of the exit condition:

If the number of the current iteration reaches the number of the predetermined maximum iteration, then exit. If else; go to step 2.

Contrary to the BP, the PSO algorithm avoids the convergence to a local minimum, since it is not founded on gradient information (Abbass et al. (2001)). The objective of the PSO is to produce the best set of weights (particle position) where numerous particles are moving to get the best solution, where the total number of weights characterize the dimension of the search space. The optimization is finished when the personal best solution of each particle and the global best amount of the entire swarm are attended.

### 2.3 The Wavelet local linear wavelet neural WLLWNN

In this section, we propose a novel neural network based wavelet decomposition; this model consists into two steps. At the first step, the historical price data has been decomposed using Wavelet decomposition and introduced, in the second step, into the network (LLWNN) to produce the set of input variables, the formed model termed WLLWNN model.

The expression (5) represents the decomposition of $X(t)$into orthogonal components at different resolutions and constitutes the so-called wavelet multiresolution analysis (MRA). In the second step, this decomposed time series is shaped through the local linear wavelet neural network. The structure of WLLWNN model is shown in Figure.1.

For the LLWNN model, wavelets are used as an activation function generated from one single function $\psi(x)$ by the operation of dilation and translation $\psi(x)$.

$$\psi(x) = \left\{ \psi_i = |a_i|^{-1/2} \psi\left( \frac{x - b_i}{a_i} \right); \quad a_i, b_i \in R^n, i \in z \right\} \tag{16}$$

$$x = (x_1, x_2, \ldots x_n)$$

$$a_i = (a_{i1}, a_{i2}, \ldots a_{in})$$

$b_i = (b_{i1}, b_{i2}, \ldots b_{in})$

$\psi(x)$ is localized in both time space and the space scale, is called a mother wavelet and the parameters $a_i$ and $b_i$ are the scale and translation parameters, respectively.

Instead of the straightforward weight $w_i$ (piecewise constant model), a linear model $\upsilon_i = w_{i0} + w_{i1}x_1 + \ldots + w_{in}x_n$ is introduced.

The activities of the linear models $\upsilon_i$ $(i = 1,2,\ldots n)$ are determined by the associated locally active wavelet functions $\psi_i(x)$ $(i = 1,2,\ldots n)$, thus $\upsilon_i$ is only locally significant. Non-linear wavelet basis functions (named wavelets) are localized in both time space and space scale. Here $m = n$ and output $(Y)$ of the proposed model is calculated as follows:

$$Y = \sum_{i=1}^{M} (w_{i0} + w_{i1}x_1 + \ldots . w_{in}x_n)\psi_i(x) \tag{17}$$

The mother wavelet is

$$\psi(x) = \frac{-x^2}{2} e^{\frac{-x^2}{\sigma^2}} \tag{18}$$

$$\psi(x) = e^{-\left(\frac{x-c}{\sigma}\right)^2} \tag{19}$$

Where $x = \sqrt{d_1^2 + d_2^2 + \ldots \ldots d_n^2}$ (20)

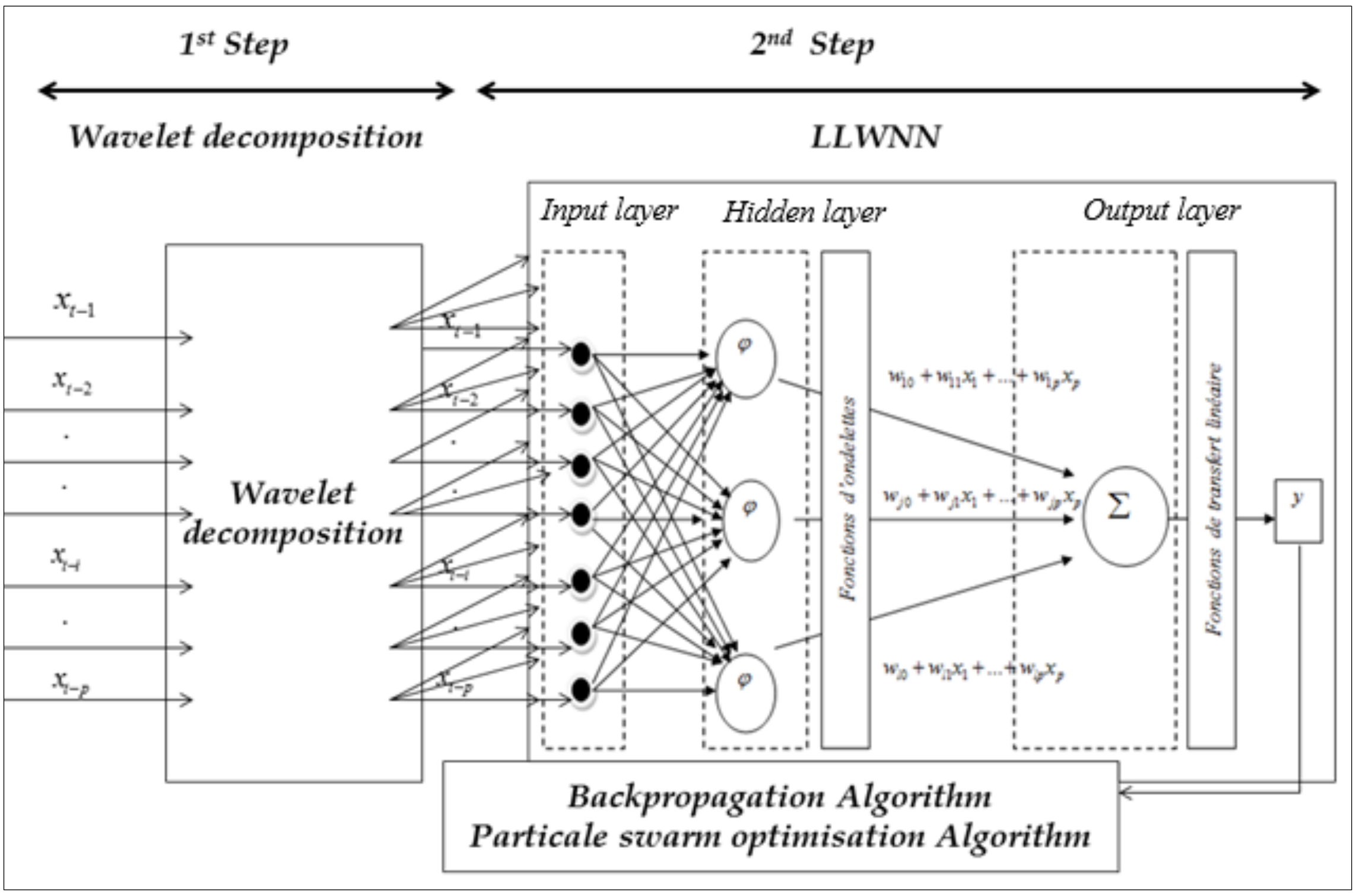

Figure 1. The proposed Wavelet Local Linear Wavelet Neural Network

### 2.4 The k-factor GARMA model

Gray et al. (1989) proposed the $k$-frequency GARMA model, which is a general from of the ARFIMA model that allow the modeling of the periodic or quasi-periodic movement in the data. This model. The generalized autoregressive k-factor GARMA model is defined as follows;

$$\Phi(L)\prod_{i=1}^{k}(I-2v_{m,i}L+L^2)^{d_{m,i}}(y_t-\mu)=\Theta(L)\varepsilon_t \tag{21}$$

Where $\Phi(L)$ and $\Theta(L)$ are the polynomials of the delay operator $L$ such that all the roots of $\Phi(z)$ and $\Theta(z)$ lie outside the unit circle. The parameters $v_{m,i}$ provide information about periodic movement in the conditional mean, $\varepsilon_t$ is a white noise perturbation sequence with variance $\sigma_\varepsilon^2$, $k$ is a finite integer, $|v_{m,i}|<1$, $i=1,2,\ldots k$, $d_{m,i}$ are long memory parameters of the conditional mean indicating how slowly the autocorrelations are damped, $\mu$ is the mean of the process, $\lambda_{m,i}=\cos^{-1}(v_{m,i})$, $i=1,2,\ldots k$, denote the Gegenbauer frequencies (G-frequencies).

The GARMA model with $k$-frequency is stationary when $|v_{m,i}|<1$, and $d_{m,i}<1/2$ or when $|v_{m,i}|=1$ and $d_{m,i}<1/4$, the model exhibits a long memory when $d_{m,i}>0$.

The main characteristic of model is given by the presence of the Gegenbauer polynomial;

$$P_m(L)=\prod_{i=0}^{k}(I-2v_{m,i}L-L^2)^{d_{m,j}} \tag{22}$$

This polynomial may be considered a generalized long-memory filter that models the long-memory periodic behavior at $k+1$ different frequencies. When thinking of the $\lambda_i$ as the driving frequencies of a cyclical pattern of length S, $\lambda_{m,i}=(2\pi i/S)$, and $k+1=[S/2]+1$, where $[\cdot]$ stands for the integer part. To highlight the contribution of $P_m(L)$ at frequencies $\lambda=0$ and $\lambda=\pi$ , equation (22) can be written as:

$$P_m(L)=(I-L)^{d_{m,0}}(I+L)^{d_{m,k}I(E)}\prod_{i=1}^{k+1}(I-2v_{m,i}L-L^2)^{d_{m,j}} \tag{23}$$

Where $I(E)=1$ if $S$ is even and zero otherwise and $k+1=[S/2]+1-I(E)$.

For a GARMA model with a single frequency, when $v=1$, the model is reduced to an ARFIMA$(p,d,q)$ model, and when $v=1$ and $d=1/2$, the process is an ARIMA model. Finally, when $d=0$, we get a stationary ARMA model.

Cheung (1993) determines the spectral density function and shows that for $d > 0$ the spectral density function has a pole at $\lambda = \cos^{-1}(\nu)$, which varies in the interval $[0, \pi]$It is important to note that when $|\nu| < 1$.

### 1.1 The Gegenbauer-GARCH model

The generalized long memory GARCH or Gegenbauer-GARCH (G-GARCH) introduced by Bordignon et al. (2007, 2010), generalized the FIGARCH and the Fractionally Integrated Exponential GARCH (FIEGARCH) models by allowing the periodic long memory patterns of the power spectrum in both the zero and non-zero frequencies and then its able to estimate the seasonality in different frequencies in the conditional variance.

The G-GARCH model for conditional variance modeling is given by;

$$\sigma_t^2 = \gamma + \beta(L)\sigma_t^2 + \left\{ I - \beta(L) - \left[ (I-L)^{d_{v,0}} (I+L)^{d_{v,k} I(E)} \prod_{i=1}^{k-1} (I - 2\nu_{v,i} L + L^2)^{d_{v,i}} \right] \psi(L) \right\} \varepsilon_t^2 \quad (24)$$

This implies that in the G-GARCH framework each frequency has been modelled by means of a specific long-memory parameter $d_{v,i}$(differencing parameter of the conditional variance). When $d_{v,0} = d_{v,1} = \ldots = d_{v,k}$, all the involved frequencies have the same degree of memory. Bordignon et al. (2007) proposed to model the logarithm of the conditional variances in order to guarantee the conditional variance positivity. For this purpose, a filter to a generalized log-GARCH model is proposed;

$$P_v(L)\psi(L)\left[\ln(\varepsilon_t^2) - \tau\right] = \gamma + [I - \beta(L)]\vartheta_t \quad (25)$$

Where $P_v(L)$is the generalized long memory filter introduced into a GARCH structure, $\vartheta_t = \ln(\varepsilon_t^2) - \tau - \ln(\sigma_t^2)$ is a martingale difference and $\tau = E[(\ln(z_t^2))]$. The expected $\tau$ value depends on the distribution of the idiosyncratic shock and ensures that $\vartheta_t$ is a martingale difference, given that $\ln(\varepsilon_t^2) = \ln(\sigma_t^2) + \ln(z_t^2)$. Under the Gaussian assumption $\tau = -1.27$.

The expression for conditional variance implied by (24) is:

$$\ln(\sigma_t^2) = \gamma + \beta(L)\ln(\sigma_t^2) + [I - \beta(L) - P_v(L)\psi(L)]\left[\ln(\varepsilon_t^2) - \tau\right] \quad (26)$$

Since we are modelling $\ln(\sigma_t^2)$ instead of $\sigma_t^2$, no constraints for variance positivity are necessary.

The $k$-factor GARMA model extended using the Gegenbauer GARCH model form the propose the dual generalized $k$-factor GARMA-G-GARCH model that is able to capture seasonality and long memory dependence in both the conditional mean and the conditional variance (see Figure 2).

Concerning the estimation of the $k$-factor GARMA-G-GARCH model, we adopt an estimation procedure based on the wavelets following the methodology proposed by Whitcher (2004).

In this paper we focus on the discrete wavelet packet transform (DWPT), which deals with the existence of seasonalities and allows to decorrelate the spectrum of the process (see Boubaker (2015) and Guegan and Lu (2009), for more details).

## 1.2 The proposed *k*-factor GARMA-WLLWNN hybrid model

Our hybrid methodology combines a $k$-factor GARMA model and the proposed WLLWNN model. The $k$-factor GARMA model offers greater flexibility in modeling simultaneous short and long-term behavior of a seasonal time series. In addition, the choice of WLLWNN in our hybrid model is motivated by the wavelet decomposition and its local linear modeling ability.

It may be reasonable to consider time series to be composed by two components; the first one is a parametric form with unknown parameters where a parametric method seems appropriate for such processes. The second component related to the residuals, this part usually present no specific process. Hence, it is difficult to determine the appropriate model that can be deal with this part of the time series. For this reason, a non-parametric model seems appropriate for modelling the residuals. This choice is motivated by the fact that the non-parametric model can reduce modelling bias by imposing no specific model structure, rather than certain smoothness assumption, and therefore, they are particularly used when we have little information or when we want to be flexible about the underling model.

Our methodology consists into two steps; in the first step, the aim is to model the conditional mean using a $k$-factor GARMA model. However, residuals are important in forecasting time series; they may contain some information that is able to improve forecasting performance. Thus, in the second step, the residuals, resulting from the first step, will be treated according to a novel wavelet local linear wavelet neural network (WLLWNN) model (see Figure.1).

Hence, a time series can be written as:

$$y_t = \mu_t + \varepsilon_t \tag{27}$$

Where $\mu_t$ denote the conditional mean of the time series, and $\varepsilon_t$ is the residuals. The $k$-factor GARMA model is used to reproduce the conditional mean $\mu_t$ using equation (21).

In the second step, the residuals from the parametric model ($\varepsilon_t$) are used as a proxy for the corresponding volatility and modeled using the WLLWNN model.

Let $\varepsilon_t$ denote the residuals at time $t$ from the $k$-factor GARMA model, then

$$\varepsilon_t = y_t - \hat{\mu}_t \tag{28}$$

Where $\hat{\mu}_t$ is the forecast value from the estimated relationship (equation 21). Thereafter, the forecast values and the residuals of the semi-parametric modelling are the results of the first stage.

In the second stage, the aim is the modelling of the residuals using the WLLWNN with $n$ input nodes, the WLLWNN for the residuals is:

$$\varepsilon_t = f(\varepsilon_{t-1}, \varepsilon_{t-2}, \ldots \varepsilon_{t-n}) \tag{29}$$

Where each $\varepsilon_{t-i}$ is decomposed using the Wavelet Transform (equation 5), $f$ is a non-linear, non-parametric function determined by the neural network with the reference to the current state of the data, during the training of the neural network. The output layer of the network (equation 17) gives the forecasting results;

$$\hat{y}_t = \hat{\mu}_t + \hat{\varepsilon}_t \tag{30}$$

Hence, this global prediction, represent the result of forecasting both, the conditional mean and the conditional variance of the time series (see Figure 2).

To conclude, the proposed hybrid model exploits the originality and the strength of the $k$-factor GARMA model as well as the WLLWNN model to detect the different features existing in the data in order to benefit the complementary characteristics of the models from which there are composed. Thus, the proposed hybrid model can be an efficient way giving a more general and accurate method than other hybrid models.

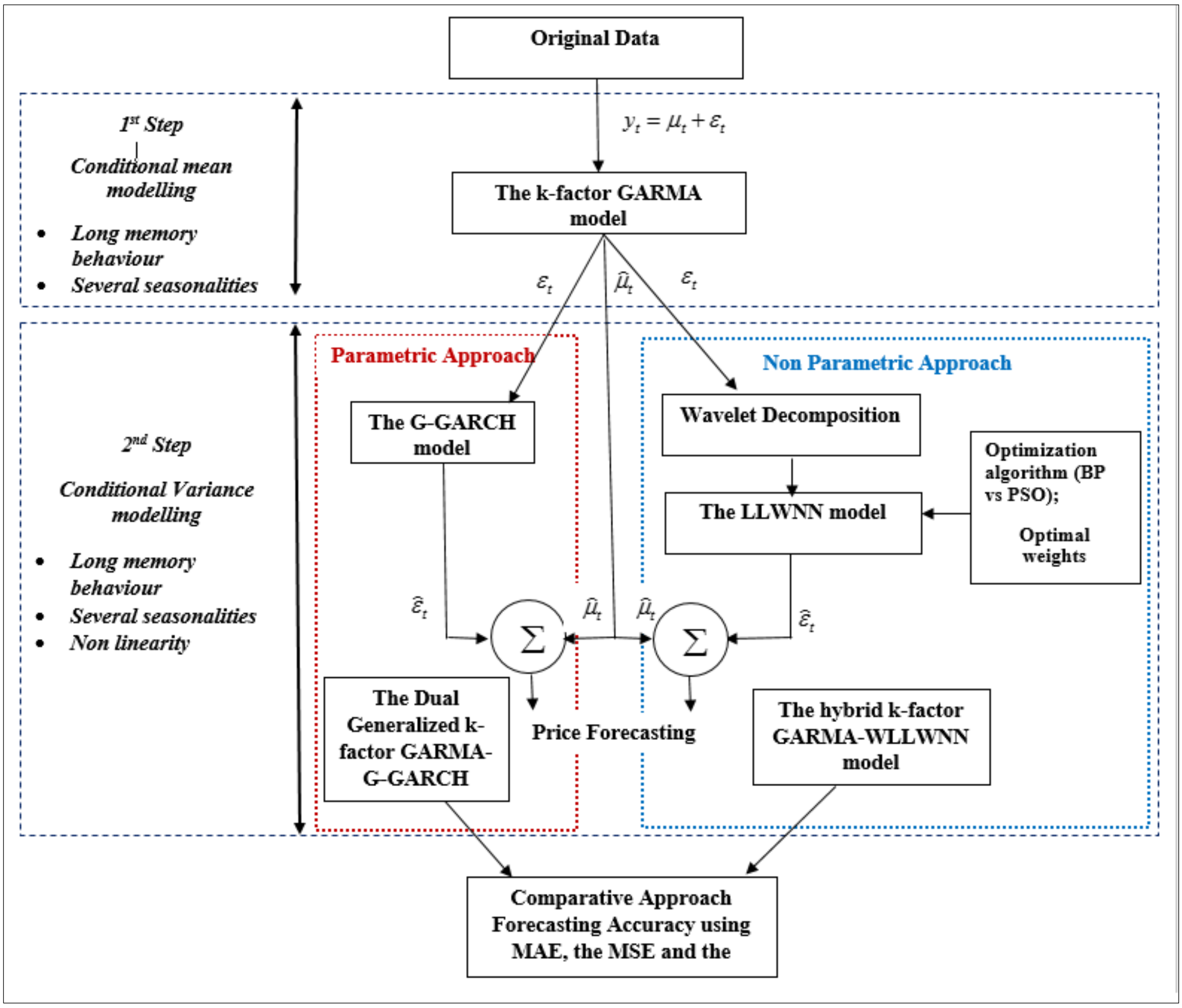

Figure 2. A schematic representation of the adopted econometric methodology

# 2. Empirical Methodology

## 2.1 Data description and Preliminary study

The predictive performance of the proposed model is tested using hourly spot prices from the Nord Pool electricity market, our sample covers the period between 1st of January 2016 and 30 June 2017, in total $T = 13128$ hourly observations, illustrated in Figure 3. In this study, we consider the logarithm of these series ( $R_t = \Delta Log P_t$ ) to make the series stationary.

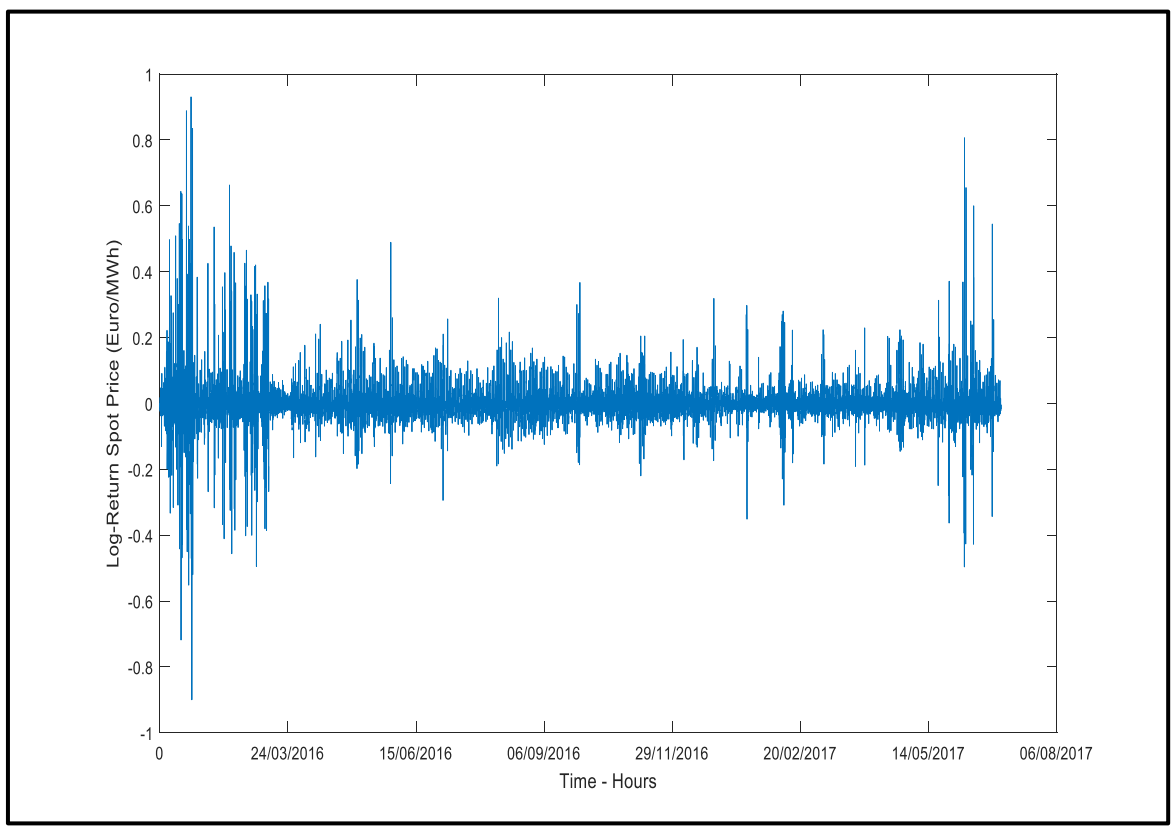

**Figure 3: Hourly Log-return of spot price for NordPool**

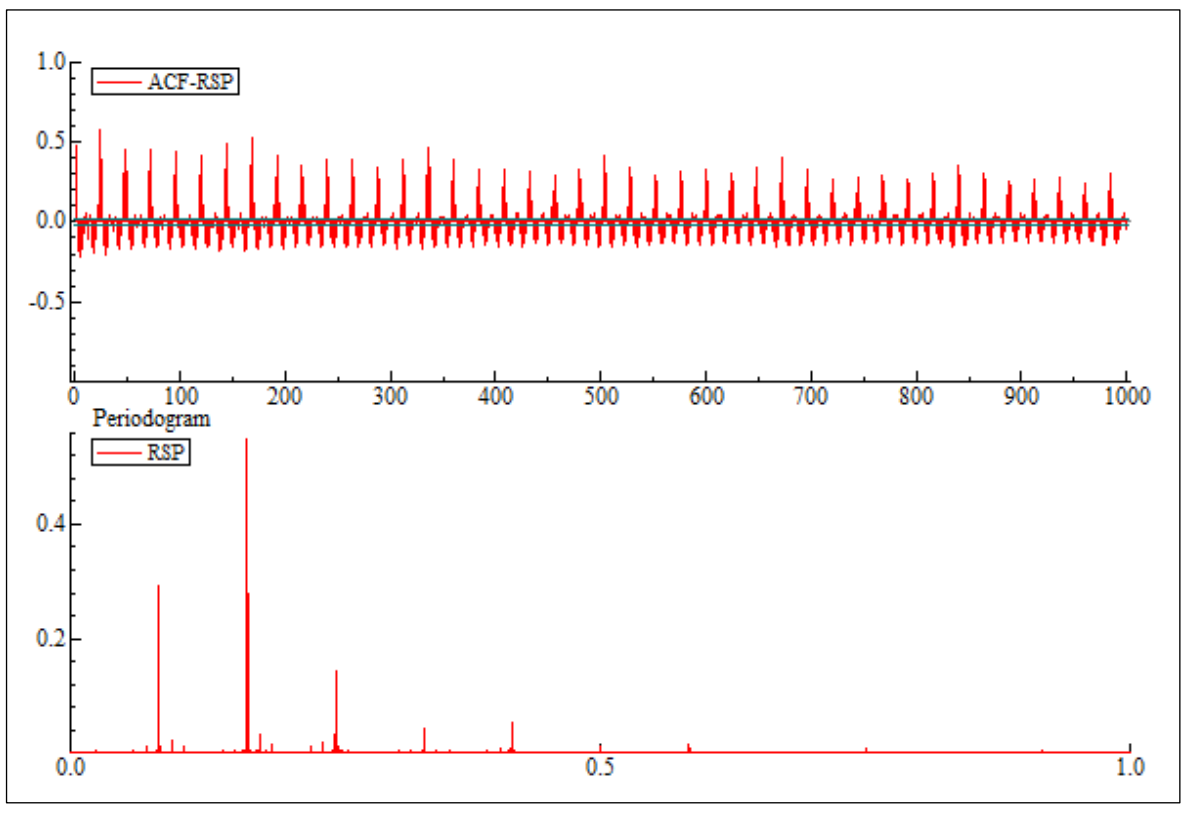

**Figure 4: Nord Pool Auto Correlation Function & Periodogram electricity market**

As shown in Figure 4, for the RSP, the spectral density, traced by the periodogram, shows several peaks at equidistant frequencies, which proves the presence of several seasonalities. Using the GPH (Geweke and Porter-Hudak, (1983)) and LW (Robinson, (1995)) statistics, we tested the long-range dependence in the conditional mean. Corresponding results shown in Table 1 indicate evidence of long memory.

**Table 1. Results of GPH and LW long-range dependence tests in the conditional mean**

| | | *GPH* | | | *LW* | | |
|---|---|---|---|---|---|---|---|
| | *Bandwidth* | $\hat{d}_m$ | *Standard error* | *p-value* | $\hat{d}_m$ | *Standard error* | *p-value* |
| *RSP* | $N^{0.5}$=115 | -0.3432*** | 0.0642 | 0.0000 | -0.4554*** | 0.0466 | 0.0000 |
| | $N^{0.6}$=296 | -0.2858*** | 0.0386 | 0.0000 | -0.3473*** | 0.0290 | 0.0000 |
| | $N^{0.7}$=764 | -0.4354*** | 0.0236 | 0.0000 | -0.6390*** | 0.0180 | 0.0000 |
| *N=13128* | $N^{0.8}$=1970 | -0.3698*** | 0.0146 | 0.0000 | -0.5210*** | 0.0112 | 0.0000 |

*Note*. Asterisks denote significance at ***1% level.

## 2.2 Estimation Results

- **Conditional mean estimation results (the $k$-factor GARMA model)**

**Table 2. Estimation of the k-factor GARMA model: a wavelet based approach**

| *Parameters* | *k-factor GARMA model estimation* |
|---|---|
| $\hat{\Phi}$ | 0.0213*** |
| $\hat{\Theta}$ | - |
| $\mu$ | - |
| $\hat{d}_{m,1}$ | 0.3892*** |
| $\hat{d}_{m,2}$ | 0.2113*** |
| $\hat{d}_{m,3}$ | 0.1048*** |
| $\hat{\lambda}_{m;1}$ | 0.0416*** |
| $\hat{\lambda}_{m,2}$ | 0.0834*** |
| $\hat{\lambda}_{m,3}$ | 0.1713*** |

*Note.* Asterisks denote significance at ***1% level.

The seasonality can be easily observed in the frequency domain $\lambda_i = 1/T$; where $\lambda$ is the frequency of the seasonality and $T$ is the period of seasonality. As shown the spectral densities, represented by periodogram (see Figure 4), are unbounded at equidistant frequencies, which proves presence of several seasonalities. They show special peaks at frequencies $\hat{\lambda}_{m,1} = 0.0416$ ($T = 24.038 \approx 24$ hours= 1 day), $\hat{\lambda}_{m,2} = 0.0834$($T = 11.99$=12 hours $\approx$1/2 day), and $\hat{\lambda}_{m,3} = 0.1713$($T = 5.83 \approx 6$ hours=1/3 day), corresponding to cycles with daily, semi-daily and third-daily periods, respectively.

The second step consist in modeling the conditional variance, so the residuals of the $k$-factor GARMA estimation are shaped through a novel WLLWNN as a first approach and then treated using the generalized GARCH model termed G-GARCH as a second approach, in order to select the adequate method.

- **Conditional variance estimation results**

  - **The WLLWNN estimation results**

In the NNs, the data training allows the model to develop a generalized structure, which is able to estimate accurately the unknown values. In order to create the training and testing data sets, it is essential to select the input variables. In reality, a number of factors like historical prices, bidding strategies, weather, fuel prices and demand elasticity, affect electricity prices. However, the process will be complicated, if we include all these factors in the forecasting model, since we need a separate module of relevant feature selection, in order to identify and include the most suitable input features. Furthermore, the redundancy in the input features can reduce the performance of the model instead of improving it (Kohavi and George (1997); and Pindoriya et al. (2008)). In order to conserve the simplicity of our methodology and for comparison purposes, we have used historical prices as the unique input variable referring to several researches.

Hence, residuals resulting from the $k$-factor GARMA modeling are considered here as the input of the novel WLLWNN in order to estimate the conditional variance. For the purpose to avoid the possibility of coupling among different input and to accelerate convergence, all the inputs are normalized within a range of [0, 1] using the following formula before applying it to the network, which considered as the most commonly used data smoothing method;

$$y_{norm} = \frac{y_{org} - y_{\min}}{y_{\max} - y_{\min}} \tag{31}$$

Where $y_{norm}$ is the normalized value, $y_{org}$ is the original value, $y_{\min}$ and $y_{\max}$ are the minimum and maximum values of the corresponding residuals data.

**Wavelet decomposition**

These normalized data are then decomposed using the MODWT[9] with Daubechies least asymmetric $(La)$ wavelet filter of length $L = 8\,(La(8))$[10]. This wavelet filter has been frequently adopted in the financial literature and it has been proved that $La(8)$ provides the best performance for the wavelet time series decomposition. Our MODWT decomposition goes up to level $J = 13$ that is specified by, $J \leq \log_2\left[\frac{T}{L-1}+1\right]$ i.e. Where $T$ represent the length of the given time series and $L$ denote the length of the filter (Percival and Walden (2000); and Gencay et al. (2002)) plotted in Figures 5 and 6. The time series is decomposed into 13 details $(D_t(1), \ldots D_t(13))$. It has been proved empirically that the level of decomposition 10 of price series (Figure 5) is optimum for an accurate forecasting. Figure 6 represents the wavelet smooth S14, it is noticed that the deterministic trend is nonlinear.

[9]The wavelet transform adapts itself intelligently to capture patterns through a wide range of frequencies and hence, has the capacity to detect events that are local in time. Therefore, the wavelet transform are considered as a perfect tool to analyze non-stationary time series. For the purpose of measuring the local regularity of a signal, it is not necessary to adopt a wavelet with a narrow frequency support, but vanishing moments are essential. If the wavelet has M vanishing moments, then we prove that the wavelet transform can be interpreted as a multiscale differential operator of order M, i.e. The first M moments of the wavelet coefficients are zero: $\int_{-\infty}^{+\infty} t^r \psi(t)dt = 0, \quad r = 1, \ldots\ldots, M.$ This gives a first relation between the differentiability of the time series and its wavelet transform decay at fine scales. According to Gencay et al. (2002) the Daubechies wavelet filter guarantees the resulting DWT coefficients be stationary and hence, protect ourselves from problems caused by a non-stationary time series.

[10] As there are no universal selection criteria of the type of wavelets and their width, this choice should be imposed by the objective in order to balance two considerations. Firstly, wavelet filters of too short width can present undesirable artifacts into the multi-resolution analysis. Since the width of the wavelet filter increases, it can better match to the features of the time series. Secondly, the impact of boundary conditions becomes more severe and the localization of MODWT coefficients decreases. For more details concerning this wavelet filter see Daubechies (1992) and Gencay et al. (2002).

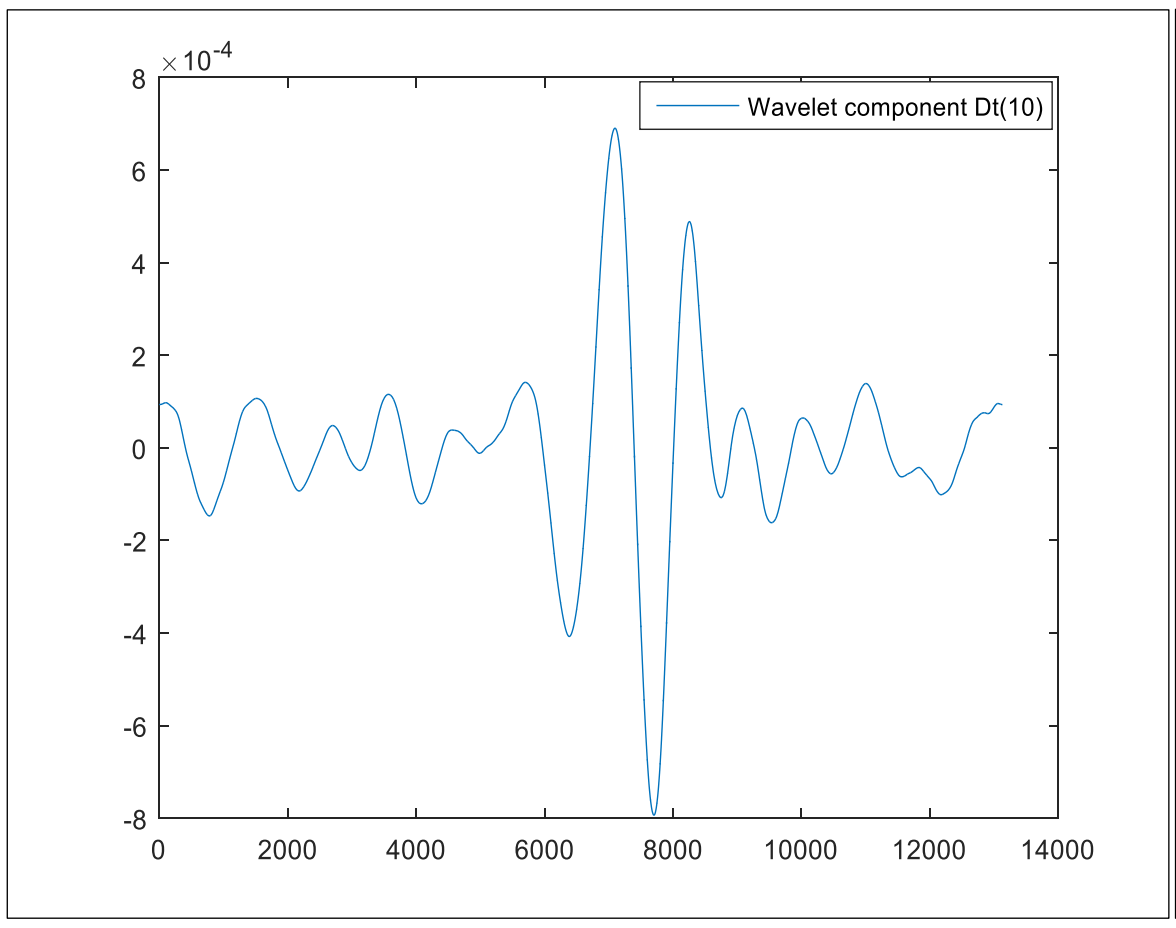

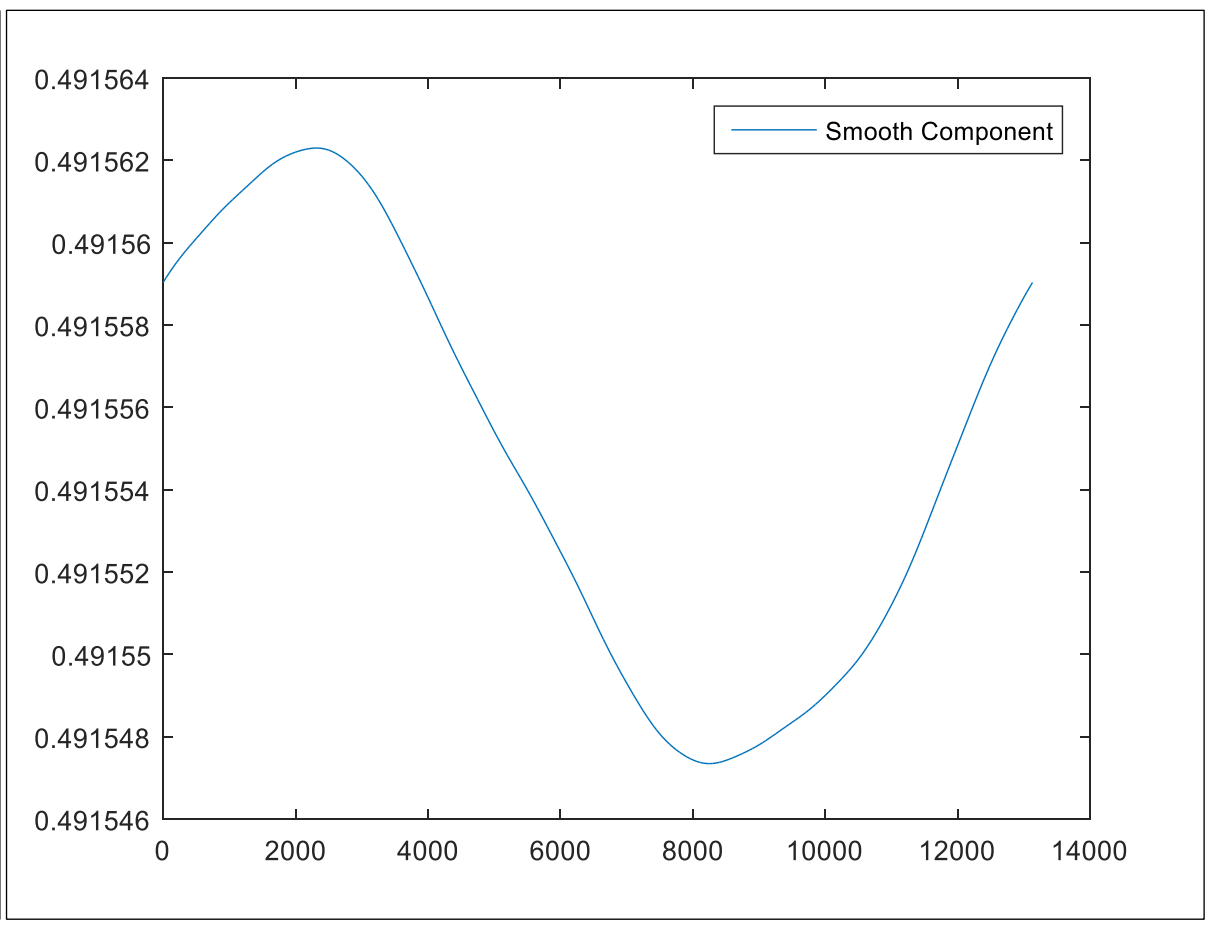

**Figure 5. Wavelet Component $D_t(10)$**

**Figure 6. Smooth Component S14**

- **The LLWNN optimization**

The dataset is divided into three successive parts as follows: (a) A sample of 2000 observations to initialize the network training, (b) a training set and (c) a test set .the forecasting experiment is performed over the test set using an iterative forecasting scheme, the models are tested for five time horizons; 6, 12, 24, 48 and 72 hours ahead. Details of the datasets are given in the Table 3.

**Table 3. Details of datasets**

| Total sample | Initialization Sample | Training Sample | Testing Sample |
|---|---|---|---|
| $N = 13128$ | From 01/01/2016 To 24/03/2016; 2000 observations | From 25/03/2016 To 27/06/2017; 11056 observations. | From 28/06/2017 to 30/06/2017; 72 observations (divided into 5 time horizons) |

In order to find the best neural network architecture the Back Propagation algorithm (BP) and the Particle Swarm Optimization Algorithm (PSO) are applied to optimize the parameters. Table 4 and Table 5 provide the summary of information related to the network architecture.

**Table 4. LLWNN based BP Algorithm architecture**

| | |
|---|---|
| *Number of hidden layers* | 10 |
| *Learning rate* | 0.5 |
| *Layer conversion function* | Wavelet Function |
| *Algorithm* | Back Propagation (BP) Learning Algorithm |

Parameters adopted for running PSO are presented as follow;

**Table 5. LLWNN based PSO Algorithm architecture**

| | |
|---|---|
| *Number of population* | 20 |
| *Number of generation* | 200 |
| $C_1, C_2$ | 1.05 |
| *Maximum velocity* | 1 |
| *Minimum velocity* | 0.3 |
| *Number of hidden layers* | 10 |
| *Learning rate* | 0.5 |
| *Layer Activation function* | Wavelet Function |

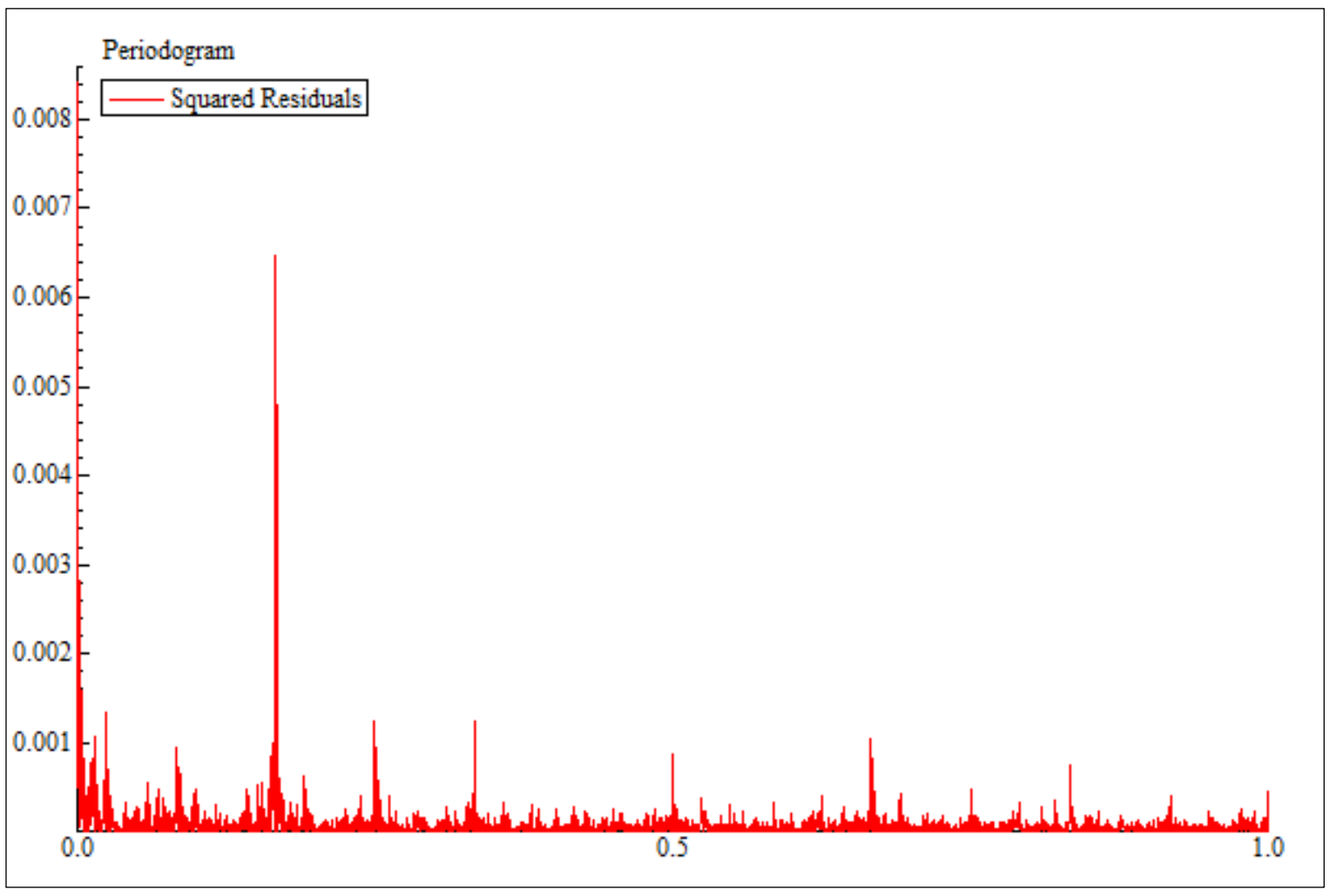

**Figure 7: Periodogram of the residuals of the $k$-factor GARMA model**

- **The $k$-factor GARMA-G-GARCH estimation results**

As shown in Figure 7, the presence of several peaks at equidistance frequencies in the spectral densities proves the presence of several seasonalities. Note that the residuals from the $k$-factor GARMA model are used as a proxy of the corresponding volatility, on which we performed Long memory tests. As reported in Table 6, the results of the GPH and LW indicate the presence of long memory in the conditional variance.

**Table 6. Results of GPH and LW long-range dependence tests in the conditional variance**

| | *Bandwidth* | *GPH* | | | *LW* | | |
|---|---|---|---|---|---|---|---|
| | | $\hat{d}_v$ | *Standard error* | *p-value* | $\hat{d}_v$ | *Standard error* | *p-value* |
| *RSP* N=13128 | $N^{0.5}$=115 | 0.4276*** | 0.0642 | 0.0000 | 0.3440*** | 0.0466 | 0.0000 |
| | $N^{0.6}$=296 | 0.4171*** | 0.0386 | 0.0000 | 0.3491*** | 0.0290 | 0.0000 |
| | $N^{0.7}$=764 | 0.2979*** | 0.0236 | 0.0000 | 0.2623*** | 0.0180 | 0.0000 |
| | $N^{0.8}$ =1970 | 0.2035*** | 0.0146 | 0.0000 | 0.1757*** | 0.0113 | 0.0000 |

*Note.* Asterisks denote significance at ***1% level.

**Table 7. The $k$ -factor GARMA-G-GARCH Estimation Results**

| *k-factor GARMA model estimation* | | *The G-GARCH model estimation* | |
|---|---|---|---|
| $\hat{\Phi}$ | 0.0217*** | $\hat{\psi}$ | 0.1315*** |
| $\hat{\Theta}$ | - | $\hat{\beta}$ | 0.1758*** |
| $\mu$ | - | $\hat{\gamma}$ | 0.0001*** |
| $\hat{d}_{m,1}$ | 0.3889*** | $\hat{d}_{v,1}$ | 0.2741*** |
| $\hat{d}_{m,2}$ | 0.2167*** | $\hat{d}_{v,2}$ | 0.1348*** |
| $\hat{d}_{m,3}$ | 0.1113*** | $\hat{d}_{v,3}$ | 0.0869*** |
| $\hat{\lambda}_{m;1}$ | 0.0418*** | $\hat{\lambda}_{v;1}$ | 0.0187*** |
| $\hat{\lambda}_{m,2}$ | 0.0832*** | $\hat{\lambda}_{v,2}$ | 0.0413*** |
| $\hat{\lambda}_{m,3}$ | 0.1717*** | $\hat{\lambda}_{v,3}$ | 0.0982*** |

*Note.* Asterisks denote significance at ***1% level.

The residuals from the $k$-factor GARMA are modeled using the G-GARCH model in order to estimate the seasonal long memory behaviour in the conditional variance. The estimation results of the $k$ -factor GARMA-G-GARCH model are reported in table 7. The spectral density, represented by periodogram (Figure 7), is unbounded at equidistant frequencies, which proves the presence of several seasonalities. They show special peaks at frequencies $\lambda_{v,1} = 0.0187$ (T=53h, 28min $\approx$2days), $\lambda_{v,2} = 0.0413$ (T=24h, 12min$\approx$1 day), and $\lambda_{v,3} = 0.0982$ (T=10h, 10min $\approx$1/2 day), that corresponding to cycles with two days, daily and semi-daily periods, respectively.

## 2.3 Forecasting Results comparaison

This section is devoted to the evaluation of the estimated models in a multi-step-ahead forecasting task. Since forecasting is basically an out-of-sample problem, we prefer to apply out-of-sample criteria. Accordingly, five different periods (6 hours, 12 hours, one day, two days, and three days) were selected in aim to ensure the robustness of forecasting results. In order to evaluate the forecasting accuracy, we apply three evaluation criteria, namely the Mean Absolute error (MAE), the Mean Squared Error (MSE) and the Root Mean Squared Error (RMSE), given respectively by:

$$MAE = \frac{1}{N - t_1} \sum_{t=t_1}^{N} \left| \left( y_{t+h} - \hat{y}_{t,t+h} \right) \right| \quad (32)$$

$$MSE = \frac{1}{N - t_1} \sum_{t=t_1}^{N} \left( y_{t+h} - \hat{y}_{t,t+h} \right)^2$$

(33)

$$RMSE = \left( \frac{1}{N - t_1} \sum_{t=t_1}^{N} \left( y_{t+h} - \hat{y}_{t,t+h} \right)^2 \right)^{1/2} \quad (34)$$

.

Where $N$ is the number of observations, $N - t_1$ is the number of observations for predictive performance, $y_{t+h}$ is the the log-return series through period $t + h$, $\hat{y}_{t,t+h}$ is the predictive log-return series of the predictive horizon $h$ at time $t$.

**Table 8. Out of sample Forecasts Results**

| *Models* | *Criterion* | $h = 6$ | $h = 12$ | $h = 24$ | $h = 48$ | $h = 72$ |
|---|---|---|---|---|---|---|
| *LLWNN based BP Algorithm* | *MAE* | 0.0078 | 0.0104 | 0.0169 | 0.0158 | 0.0101 |
| | *MSE* | $7.047 \times 10^{-5}$ | $1.4995 \times 10^{-4}$ | $5.0542 \times 10^{-4}$ | $3.6611 \times 10^{-4}$ | $1.6566 \times 10^{-4}$ |
| | *RMSE* | 0.0084 | 0.0122 | 0.0225 | 0.0191 | 0.0129 |
| *LLWNN based PSO Algorithm* | *MAE* | 0.0082 | 0.0116 | 0.0135 | 0.0137 | 0.0130 |
| | *MSE* | $1.0331 \times 10^{-4}$ | $2.4644 \times 10^{-4}$ | $2.6787 \times 10^{-4}$ | $3.1191 \times 10^{-4}$ | $2.9709 \times 10^{-4}$ |
| | *RMSE* | 0.0102 | 0.0157 | 0.0164 | 0.0177 | 0.0172 |
| *The hybrid k-factor GARMA-LLWNN based BP algorithm* | *MAE* | 0.0038 | 0.0081 | 0.0087 | 0.0101 | 0.0089 |
| | *MSE* | $3.1983 \times 10^{-5}$ | $9.5044 \times 10^{-5}$ | $1.2492 \times 10^{-4}$ | $1.6142 \times 10^{-4}$ | $1.2499 \times 10^{-4}$ |
| | *RMSE* | 0.0057 | 0.0097 | 0.0112 | 0.0127 | 0.0112 |
| *The hybrid k-factor GARMA-LLWNN based PSO algorithm* | *MAE*<br>*MSE*<br>*RMSE* | 0.0016<br>$3.5028\times10^{-6}$<br>0.0018 | 0.0023<br>$3.6221\times10^{-6}$<br>0.0019 | 0.0057<br>$2.2690\times10^{-5}$<br>0.0047 | 0.0061<br>$2.5612\times10^{-5}$<br>0.0050 | 0.0076<br>$3.9241\times10^{-5}$<br>0.0062 |
| *WLLWNN based BP Algorithm* | *MAE*<br>*MSE*<br>*RMSE* | $4.6968\times10^{-6}$<br>$2.2306\times10^{-11}$<br>$4.7229\times10^{-6}$ | $5.1739\times10^{-6}$<br>$2.7097\times10^{-11}$<br>$5.2055\times10^{-6}$ | $4.0551\times10^{-6}$<br>$1.7424\times10^{-11}$<br>$4.1741\times10^{-6}$ | $1.8927\times10^{-6}$<br>$4.4187\times10^{-12}$<br>$2.1021\times10^{-6}$ | $4.5983\times10^{-6}$<br>$2.6346\times10^{-11}$<br>$5.1329\times10^{-6}$ |
| *WLLWNN based PSO Algorithm* | *MAE*<br>*MSE*<br>*RMSE* | $3.7001\times10^{-8}$<br>$1.7781\times10^{-15}$<br>$4.2170\times10^{-8}$ | $7.2902\times10^{-8}$<br>$6.7075\times10^{-15}$<br>$8.1899\times10^{-8}$ | $1.6286\times10^{-7}$<br>$3.3395\times10^{-14}$<br>$1.8274\times10^{-7}$ | $5.9591\times10^{-8}$<br>$7.7661\times10^{-15}$<br>$8.8121\times10^{-8}$ | $2.7986\times10^{-7}$<br>$9.8627\times10^{-14}$<br>$3.1405\times10^{-7}$ |
| *The hybrid k-factor GARMA-WLLWNN based BP Algorithm* | *MAE*<br>*MSE*<br>*RMSE* | $2.0409\times10^{-6}$<br>$4.3281\times10^{-12}$<br>$2.0804\times10^{-6}$ | $1.7647\times10^{-6}$<br>$3.3223\times10^{-12}$<br>$1.8227\times10^{-6}$ | $2.7731\times10^{-6}$<br>$8.0457\times10^{-12}$<br>$2.8365\times10^{-6}$ | $3.6631\times10^{-6}$<br>$1.3885\times10^{-11}$<br>$3.7263\times10^{-6}$ | $6.4121\times10^{-7}$<br>$4.2383\times10^{-13}$<br>$6.5102\times10^{-7}$ |
| *The hybrid k-factor GARMA-WLLWNN based PSO Algorithm* | *MAE*<br>*MSE*<br>*RMSE* | $4.7726\times10^{-9}$<br>$3.6761\times10^{-17}$<br>$6.0631\times10^{-9}$ | $2.9456\times10^{-8}$<br>$1.0776\times10^{-15}$<br>$3.2826\times10^{-8}$ | $3.1504\times10^{-8}$<br>$1.8611\times10^{-15}$<br>$4.3140\times10^{-8}$ | $2.7194\times10^{-8}$<br>$1.2513\times10^{-15}$<br>$3.5374\times10^{-8}$ | $2.9496\times10^{-8}$<br>$1.4299\times10^{-15}$<br>$3.7813\times10^{-8}$ |
| *The k-factor GARMA-G-GARCH model* | *MAE*<br>*MSE*<br>*RMSE* | $5.2316\times10^{-7}$<br>$3.2458\times10^{-16}$<br>$1.8000\times10^{-8}$ | $6.7428\times10^{-7}$<br>$8.5214\times10^{-14}$<br>$2.9191\times10^{-7}$ | $4.8637\times10^{-6}$<br>$2.9452\times10^{-14}$<br>$1.7160\times10^{-7}$ | $3.6238\times10^{-8}$<br>$4.5831\times10^{-15}$<br>$6.7700\times10^{-8}$ | $3.2478\times10^{-8}$<br>$2.9773\times10^{-15}$<br>$5.4600\times10^{-8}$ |

In order to evaluate the prediction performance of the proposed hybrid methodology, this paper has taken into account five models: the individual LLWNN model, the hybrid $k$-factor GARMA-LLWNN model, the proposed WLLWNN model, the hybrid $k$-factor-GARMA-WLLWNN model and the $k$-factor GARMA-G-GARCH model, we applied two different learning algorithms (BP and PSO) for the training of the networks. Moreover, we adopted five time horizons; 6 hours, 12 hours, one day, 2 days and 3 days ahead forecasting, using the MAE, the MSE and the RMSE out of sample criteria. The forecast evaluation results are reported in Table 8. They show that:

- The individual LLWNN based PSO algorithm outperforms the individual LLWNN based BP algorithm, in addition, the individual WLLWNN based PSO algorithm outperforms the individual WLLWNN based BP algorithm; these results prove the superiority of the PSO algorithm for training neural network model. This result can be explained by the fact that in the case of the BP algorithms weights are updated in the direction of the negative gradient. Hence, the network training with BP algorithms presents some drawbacks such as very slow convergence to a local minimum. However, in the case of training with PSO algorithm, weights are characterized by particle position. These particles velocity and position are updated, in order to search for personal best and global best values. This will avoid the convergence of weights to a local minimum.

- The hybrid $k$-factor GARMA-LLWNN model outperform the individual LLWNN model, hence, the individual LLWNN model is unable to detect, to model and to predict the features existing in the electricity prices, meanwhile, when it is compared with the hybrid model, this last one provides prediction that is more accurate. Consequently, this network need an external filter to better estimate the data since the adoption of the $k$-factor GARMA model in the first step enhance the results of forecasting. This result contradicts existing literature; Pany (2011), Pany and Ghoshal (2013) and Chakravarty et al. (2012) proving that the use of individual LLWNN models for forecasting electricity prices provides the best prediction results compared to other ordinary networks such as multi-layer perceptron networks (MLP). This is because the features of the electricity prices can be detected by the wavelet basis activation function for hidden layer neurons without any external decomposer/ composer and not having too many hidden units. In addition, this result confirms the effectiveness of the hybrid model in improving the forecasting accuracy.
- The $k$-factor GARMA-G-GARCH model outperforms the hybrid $k$-factor GARMA-LLWNN model in terms of prediction accuracy. This can be explained by the fact that the $k$-factor GARMA-G-GARCH model takes into account the seasonal long-memory in both the conditional mean and the conditional variance, making this model a robust tool that can deal with the features of the electricity prices. Furthermore, despite its capacity as a nonlinear, nonparametric model, and its particularity by having a wavelet activation function and local linearity, the LLWNN model is unable to predict the seasonality and the long memory dependence in the data. Since, when it is compared with the G-GRACH model, this last one provides a prediction results that is more accurate. This is explained by the ability of the G-GARCH in modelling the seasonal long memory behaviour in the conditional variance. This also proves the importance of taking into account the seasonal long memory behaviour to enhance forecasting accuracy (Nowotarski and Weron (2016)).
- The novel WLLWNN overcomes the limitation of the LLWNN that is related to the inability of the network to detect and model the periodic long memory behaviour in the data, since it shows its effectiveness when we compare it with the $k$-factor GARMA-G-GARCH and the $k$-factor GARMA-LLWNN model.

- As shown in Figures 8 the predictions of the $k$-factor GARMA-WLLWNN model based PSO algorithm for all the five horizons are very close to the real values (see Figure.19). That is confirm the forecasting results (Table 10), which indicate that the $k$-factor GARMA-WLLWNN model prediction errors are the smallest for all evaluation criteria and for all forecast time horizons.

To sum up, the hybrid $k$-factor GARMA-WLLWNN model outperforms all other computing techniques. In fact, this model uses the strength of three techniques at the same time; firstly, the semi-parametric $k$-factor GARMA model that allows detecting and estimating both the long memory and the seasonality in the conditional mean. Secondly, the wavelet decomposition, which can produce a good local representation of the signal in both time and frequency domains and hence it's a good tool to bring out the hidden patterns in the electricity prices, such as high volatility, corrupted by occasional spikes and follows by several seasonalities. Finally, with the capacity of the LLWNN model as a nonlinear, nonparametric model, and its particularity by having a wavelet activation function and local linearity, this network can capture more subtle aspects of the data. Hence, the proposed hybrid $k$-factor GARMA-WLLWNN is a robust tool that can be dealing with the features of the electricity prices and provide the best forecasting results.

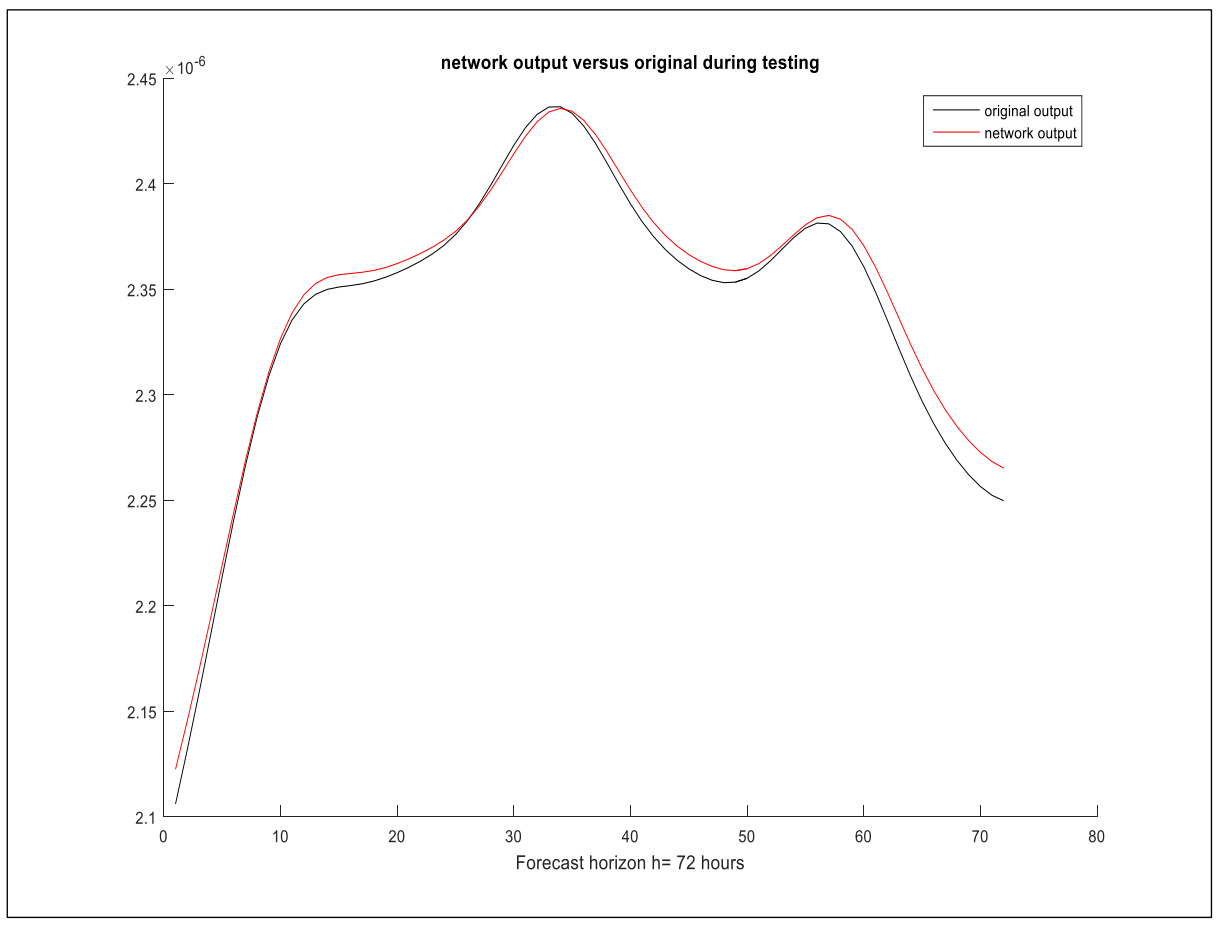

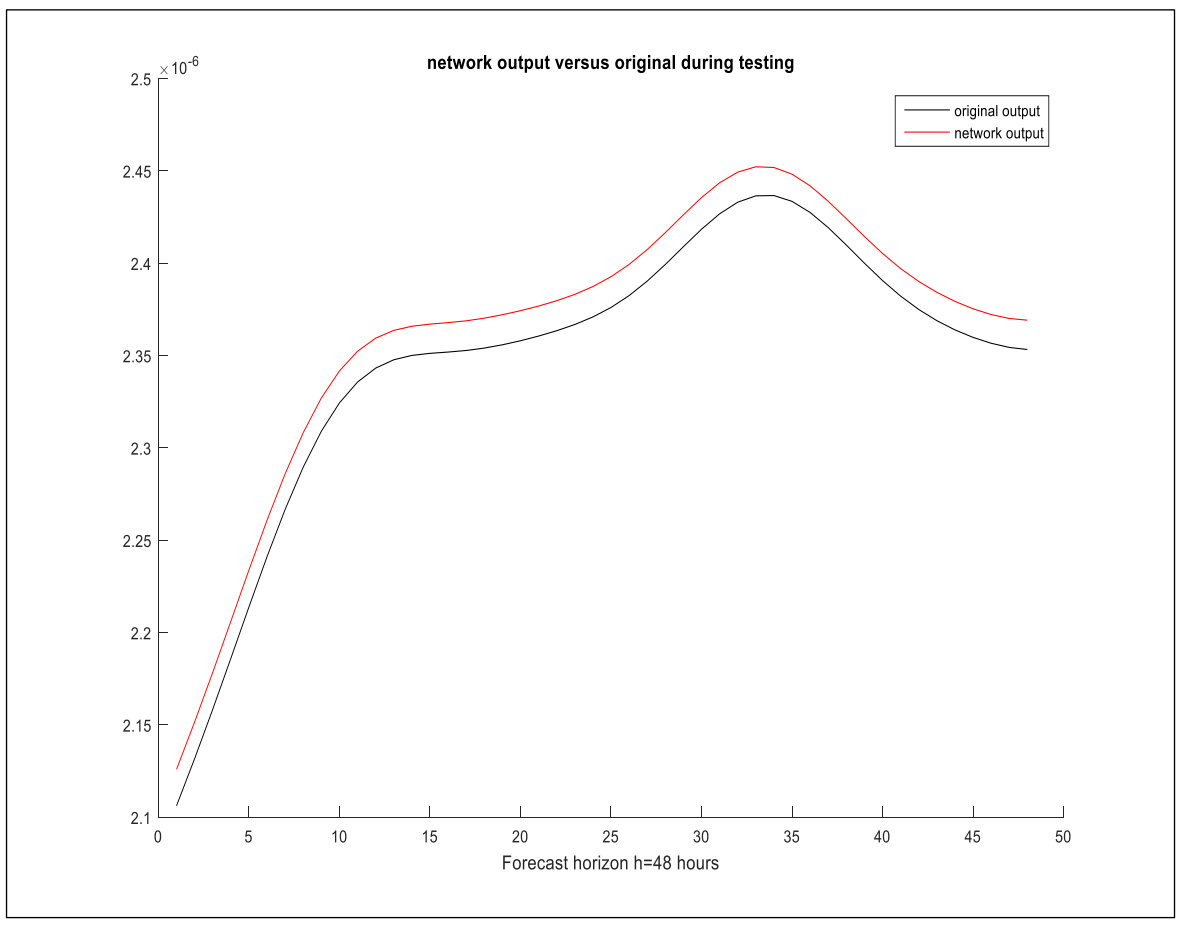

(a) (b)

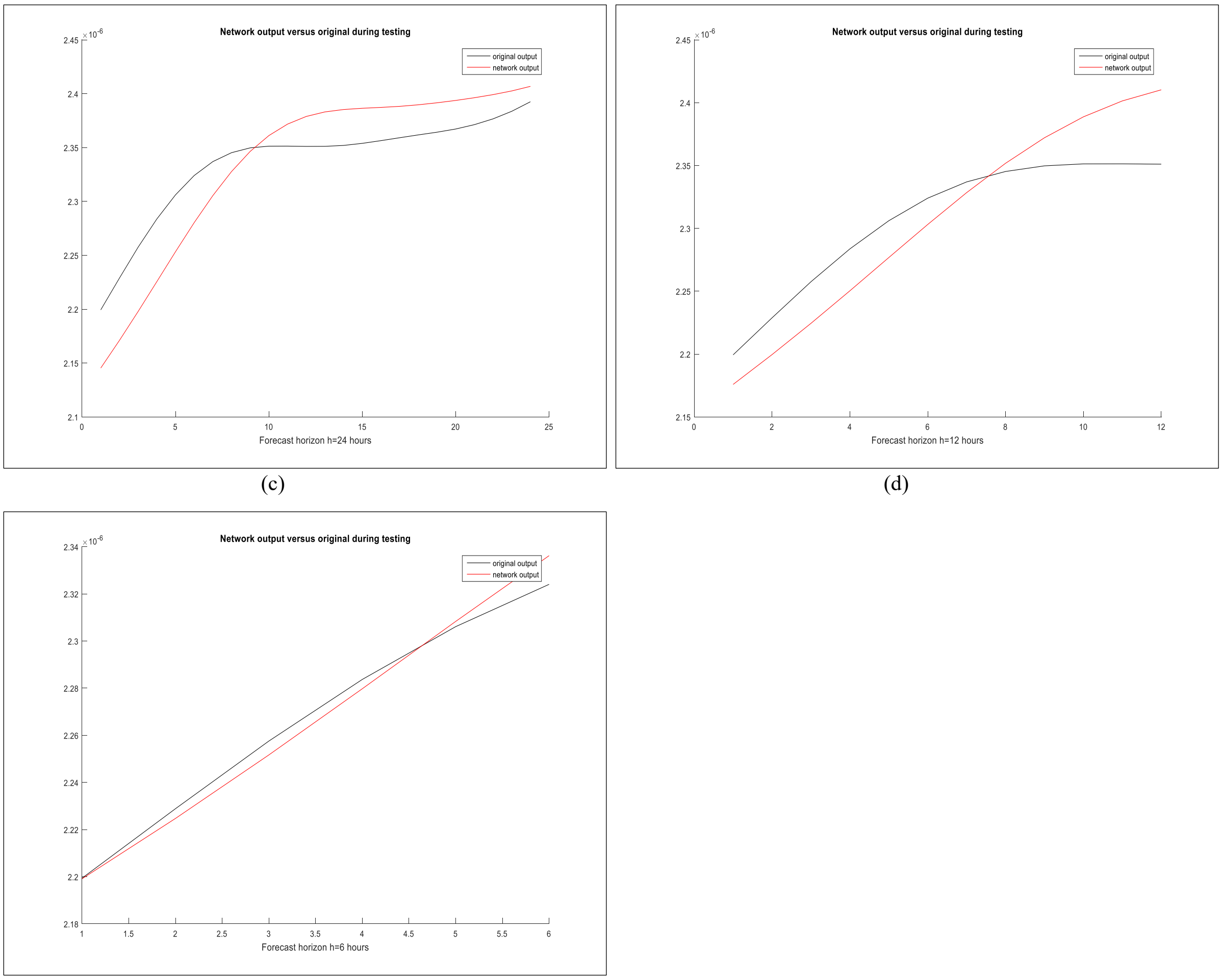

(e)

**Figure 8. Forecasting using the k-factor GARMA-WLLWNN model based PSO Algorithm;** (a) Forecasting horizon h=72, (b) h=48, (c) h=24, (d) h=12 hours, (e) h=6 hours.

## 3. Conclusion

Electricity price forecasting is crucial in the decision making of power market participants. Developing more accurate price forecasting methods has become an important research topic. In this paper, we develop a relatively novel neural network model, termed Wavelet Local Linear Wavelet Neural Network (WLLWNN). In this model, the historical price data has been decomposed into the wavelet domain constitutive sub-series using Wavelet Transform and then introduced into the network to produce the set of input variables and form the WLLWNN model. This novel network is claimed to have a higher generalization performance than the LLWNN. On the other hand, when we deal with neural networks it is very important to choose an appropriate algorithm for training, so this paper presents a comparison of two learning algorithms; the BP and PSO algorithms. The BP algorithms update weights in the direction of the negative gradient. NNs training with BP algorithms presents certain drawbacks such as very slow convergence that can be trapped in the local minimum. However, weights in PSO algorithm are represented by particle position. This particle velocity and position are updated, with the aim to search for personal best and global best values. This will avoid the convergence to a local minimum. Our experimental results showed the superiority of PSO algorithm. Moreover, the effectiveness of the novel WLLWNN is further enhanced by coupling it with the semi-parametric $k$-factor GARMA model to develop a hybrid model termed as $k$-factor GARMA-WLLWNN model. In this way, the unique strength of each method is combined to detect different patterns in the data. The performance of the proposed model is evaluated using data from Nordic Electricity market and compared with the dual generalized long memory $k$-factor GARMA-G-GARCH model, the hybrid $k$-factor GARMA-LLWNN model, and the individual WLLWNN. The empirical results prove that the proposed $k$-factor GARMA-WLLWNN method is the most suitable price forecasting technique. Since it is can produce smaller predicting errors than the other computing techniques.

To recapitulate, in addition of developing an efficient method for price forecasting, we presented a detailed analysis of several price-forecasting techniques. Our proposed hybrid model shows its performance; moreover, it is suitable and feasible to real world application since it exploits information that is available to the public (historical prices) that can be easily collected by all market participant before submitting their bids in the market and gives to investors enough time to respond.